Mark Solms,[1,2] St John Grimbly,[1,3]
Bruce Bassett,[1,3,4,5] Evert Boonstra,[1,2]
Rowan Hodson,[6] Nicolas Kuske,[1,3]
Kival Mahadew,[1,3] Benjamin Rosman,[4,5]
Charel van Hoof[7] & Jonathan Shock[1,3,8,9]

# *Inferring Affective Consciousness in an Artificial Agent*

## *A Case Study*

**Abstract:** *Creatures that display 'hedonic place preference behaviour' are thought by many scientists to experience feelings, on the assumption that their attraction to pleasure-producing substances which lack nutritional value (e.g. cocaine, morphine) cannot easily be attributed to unconscious instinctual behaviour. In this paper, we discuss how a*

Correspondence:
*Email: mark.solms@uct.ac.za*

1 Neuroscience Institute, University of Cape Town, South Africa.
2 Dept. of Psychology, University of Cape Town, South Africa.
3 Dept. of Mathematics & Applied Mathematics, University of Cape Town, South Africa.
4 School of Computer Science and Applied Mathematics, University of the Witwatersrand, Johannesburg, South Africa.
5 Machine Intelligence and Neural Discovery Institute, University of the Witwatersrand, Johannesburg, South Africa.
6 Laureate Institute for Brain Research, University of Tulsa, OK, USA.
7 Dept. of Cognitive Robotics, Delft University of Technology, Delft, Netherlands.
8 National Institute for Theoretical & Computational Sciences, Stellenbosch, South Africa.
9 Institut National de la recherche scientifique, Montreal, Quebec, Canada.

*simple artificial agent that instantiates attributes of an affective system engaging in felt uncertainty about its intrinsic needs in relation to environmental resources can similarly display hedonic place preference behaviour — through an apparently subjective form of information processing — while simultaneously being entirely deterministic. We outline some implications of this artificially engineered behaviour for our understanding of the physical basis of consciousness and the experience of free will.*

## 1. Introduction

We will use the term 'consciousness' in what has become a widely accepted sense: 'An organism has conscious mental states if and only if there is something it is like to *be* that organism — something it is like *for* the organism' (Nagel, 1974). What appeals to us about this working definition is that it licenses us to reduce notoriously slippery terms like 'subjectivity' and 'qualia' to their simplest possible meanings: sentient 'being' and 'something-it-is-like-ness' respectively. We will use these terms in these deflationary senses in this paper.

Nagel wrote also: 'If we acknowledge that a physical theory of mind must account for the subjective character of experience, we must admit that no presently available conception gives us a clue about how this could be done' (*ibid.*). He added: 'It seems unlikely that any physical theory of mind can be contemplated until more thought has been given to the general problem of subjective and objective.' In this paper we give some thought to the problem of subjective and objective in the hope of providing a clue as to how a physical theory of mind might account for the felt character of experience. We give some thought in the hope of providing *a clue*; nothing more ambitious than that.

We are currently investigating what it might take for an artificial intelligence (AI) to have conscious mental states in the sense just quoted, and we are asking how that might come about. In this paper, therefore, we will also address the question: how will we know (or agree) if and when the specific AI system that we are developing achieves consciousness? Our answer will not be about behaviour but about structure and function.

Focusing on a single case has advantages, because the methods that could be deployed to evaluate whether a system has conscious mental states vary widely depending upon the type of system being evaluated. Large language models (LLMs), for example, can be prompted to provide verbal reports which at first sight appear to reflect their subjective states, but which could easily be an aspect of their training to produce the responses that would be expected of such a system. Our system does not have this reporting capability. It is designed to utilize endogenous affective states in its manoeuvring through a very simplified world; it is not designed to communicate verbally. The widely used criterion of 'reportability' is therefore not appropriate for a system like ours, just as it isn't appropriate for evaluating the subjective states of non-verbal animals. In such systems, one must rely upon less direct ways of inferring subjective states.

This paper is divided into two further parts. Part 2 describes the rudiments of our AI system, some potential philosophical implications, and the scientific assumptions regarding consciousness upon which it is built. Part 3 describes our system in more detail, by considering how we might go about evaluating whether there is something it is like to be such a system, and whether it experiences 'free will'.

## 2.

Regarding our scientific assumptions and their philosophical implications, we provide only a brief summary here, since our assumptions and their implications are explicated and justified in detail elsewhere (e.g. Solms and Friston, 2018; Solms, 2019; 2021).

### *2.1. Background scientific assumptions and philosophical implications*

Our most basic assumption is that *affect*, rather than perception or cognition, is the most fundamental form of consciousness, both in the biological sense that it evolved first and in the physiological sense that it is a prerequisite for perceptual and cognitive forms of consciousness. Affect seems to coincide with what is traditionally called the 'waking state' (Zeman, 2001), notwithstanding the fact that it occurs also in dreams; it seems to be the state of consciousness that 'enables' all other conscious states (Koch, 2004). This is not an idiosyncratic view; it was articulated by Panksepp (1998) more than 25 years ago and it is shared by several other prominent neuroscientists (e.g. Merker, 2007; Damasio, 2010; Barrett, 2017; Seth, 2021; Cleeremans and Tallon-Baudry, 2022).

If affect is the fundamental form of consciousness, this might have implications for the hard problem of consciousness. Chalmers (1995) famously claimed that 'there is no cognitive function such that we can say in advance that explanation of that function will *automatically* explain experience'. Perhaps he could assert this only because he was talking about *cognitive* functions. In our view, we can say in advance that explanation of a particular affective function will automatically explain experience. This does not apply to *all* affective functions, some of which need not necessarily involve consciousness, but it does seem to apply to the affective function of *feeling* — a function which is intrinsically conscious.[10] We shall explain the functional mechanism of feeling in more detail below. We doubt that this function (as opposed to that of vision) can 'go on "in the dark", without any inner feel' (*ibid.*). Explaining feeling in mechanistic terms might therefore narrow the explanatory gap. Although it is still a psychological function to be explained, it seems to necessarily imply something like subjective experience.

Consider Locke's inverted spectrum argument, for example: if you swap one person's redness for another person's greenness, it has no causal consequences; but if you swap one person's pleasure for another's pain, this will surely lead them to make choices which preclude long-term survival.[11]

Since we are following Nagel's (1974) definition of consciousness here, we will hopefully be forgiven for appealing to his authority in passing. He shares our view that 'the most promising place to start on this quest [for a monistic account of consciousness] is with experiential affect' (Nagel, 2023). He continues as follows (using the word 'affect' where we would use 'feeling'):

> Mark Solms has recently made a persuasive case that the core of conscious experience is not perception or cognition but affect. This means that affect, with its positive or negative valence, is sufficient for consciousness, and that when other mental functions (that can also operate unconsciously) are conscious it is because they are enlisted in the service of affect. This suggests that if there is evidence for the kind of psychophysical monism that I think we should be looking for, it would appear

---

10 Affective feeling is an arousal state (high or low) that evaluates an organism's subjective condition (good or bad) in terms of qualitatively differentiated categories (overheating vs. suffocation vs. sleepiness, etc.) which are selectively prioritized (only the currently salient categories are felt).

11 We concede that this shows only that if an affect is conscious there are constraints on its phenomenal character which would not apply to purely perceptual states. For more comprehensive discussions of this subtle issue, see Campbell (2000); Mørch (2017; 2020).

> in its purest form in cases of biologically-based basic affect: bodily pleasure and pain, hunger and thirst and their satiation, for example. (*ibid.*)

Nagel (2021) accepts also that an 'affective zombie' is a logical impossibility.

We are well aware that some of the philosophical assumptions that we have just articulated are controversial, so we should clarify that the argument we want to present in this paper is not contingent upon these background assumptions. We have mentioned them here, at the outset, merely to provide context for an explanation of the functional architecture of our AI system, an understanding of which *is* essential for the issues that we wish to address.

### *2.2. Rudiments of an affective system*

The neuroscientists mentioned earlier — those who consider affect to be the fundamental form of consciousness — argue that its *functional mechanism* is an extended form of homeostasis (and allostasis). All living organisms must remain within their phenotypically viable bounds. The settling point for core body temperature in humans, for example, is 36.5–37.5 °C. Apparent deviation from this ideal demands either work, which, if effective, returns the organism to homeostasis, or a reappraisal of the source of the measurement. Effective work depends upon reliable predictions as to what should be done. Many such predictions are innate: they take the form of the reflexes and instincts that are provided by natural selection. In the case of thermoregulation, for example, an overheated mammal might perspire and pant, which reliably cools it down. However, inborn actions become ineffective in contexts that don't match phenotypically expected parameters. For example, autonomic breathing (which maintains blood-gas balance in most mammals) doesn't succeed if the animals are caught in a burning building, where the expected ratio between oxygen and carbon dioxide in the ambient air is distorted in favour of the latter gas. In such circumstances, the behaviour of the trapped population becomes more stochastic, and only the subset of individuals who happen to take the right actions will survive and reproduce, thereby passing down any genes that inclined them fortuitously to perform the effective actions, which could then become the basis for a new instinctual behaviour.

This mechanism is all well and good for the minority that survived, and for their offspring; but what about the majority that didn't? If they could have registered how well or badly they were doing, in some way, before it was too late, things might have turned out better for them. This 'some way', in our view, is the function of *feeling*.

Organisms in the above situation that are equipped with the function of feeling would have registered (in the form of valenced arousal: suffocation alarm, a.k.a. air hunger) how well or badly their exploratory actions were achieving the required outcome, before it was too late, thereby enabling them to change their minds and act differently. This simple function — felt affect, which makes the organism register decreasing carbon dioxide (for example) as 'good' and increasing carbon dioxide as 'bad' — is the formal basis of *choice* and therefore of voluntary action, which, through trial and error learning (the so-called Law of Effect; Thorndike, 1911), underwrites learning from experience. The subjective value of goodness and badness that inheres in affect is, of course, given by the objective value-system that under-pins the whole of evolution, namely that it is good to survive (and reproduce) and bad to die.

It is important to add that homeostatic needs must be prioritized. The prioritization process — i.e. determining which of an organism's many needs will be felt — depends crucially upon *context*: needs in relation to other needs, and needs in relation to opportunities. For example, although one's bladder might distend throughout the delivery of a long lecture, one feels the need to urinate only at the end of the lecture, when the opportunity for relief presents itself (but if the need becomes too acute, the feeling intrudes during the lecture).

Felt affect, being the guiding principle by which we explore situations outside of our expected or habituated environment, bestows an enormous selective advantage, which (we claim) is why it evolved in the first place, and why it has been conserved for at least 550 million years — if we recognize its presence in vertebrates only, which seems very restrictive (see Andrews *et al.*, 2024).

But why is affective arousal *experienced*?

First: the keynote of affect is valence, and valence is intrinsically *subjective* in a specific sense. Decreased availability of oxygen in a burning building is bad for an individual organism (of a particular phenotype) that happens to be caught in the building, but it is not bad for the universe. In other words, it is bad only from the viewpoint of being the organism, *for* the organism (and perhaps also for some other organisms that share most of its genes). Here we are speaking of intrinsic subjectivity in a rather limited sense; we shall have much more to say about it later via the concept of a 'Markov blanket'.

Second: in organisms with multiple categories of need — that is, in complex organisms — felt affect is intrinsically *qualitative*. This feature arises naturally from the fact that such organisms have many physiological parameters that must be regulated homeostatically, which means that each of them must be monitored in its own right. To put it simply, 8/10 of overheating is not the same thing as 8/10 of suffocation. If they were the same, they would be reducible to a common denominator: 16/20 of 'total need'. Then the organism would need only to lower the numerator — by avoiding flames in the burning building, for instance, without simultaneously bothering about ambient oxygen. In reality, if it did that, it would suffocate. Multiple organismic needs therefore cannot be reduced to a simple weighted sum of continuously graded variables;[12] each need must be treated categorically in the context of expected outcomes; and categorical variables are distinguished qualitatively. This is the defining feature of categorical variables (such as bodyweight vs. height vs. age) as opposed to continuous variables (such as bodyweight or height alone). The difference between weight and height is qualitative; the difference between 70kg and 90kg is quantitative. The intrinsically categorical nature of biological needs might well be the evolutionary basis of qualia.

Third: affect is intrinsically but *selectively* salient. If affect guides voluntary behaviour, then the different categories of need that it announces must be prioritized, since you cannot attend to everything at once. Also, there is an unavoidable bottleneck when it comes to voluntary action (relative to autonomic action). Affective prioritization therefore seems to be the mechanistic basis for selective focus, at given points in time, upon aspects of what must be done and what is being done. Unless purposefully directed, only prioritized needs (the ones the organism is uncertain about meeting, which demand voluntary action) are felt — as happens when the normally unconscious need to breathe becomes conscious — otherwise they are regulated stereotypically.

Combining these three intrinsic features of affect, we arrive 'automatically' at a functional, mechanistic account of experienced feeling: affective valence just is subjective, it just is differentiated qualitatively, and it just is selectively salient. Using these terms in their most deflationary sense, that is what a feeling *is*: it is a valenced, qualitatively distinctive, subjective state of the organism, which is selectively prioritized over unfelt states. Later — when we introduce the concept of Markov blankets — we shall learn that feeling states are also (necessarily) inferred. But there is nothing spooky about them. If Mary, of knowledge argument fame (Jackson, 1982), were an affective rather than a visual neuroscientist, she would know that a system with this functionality must infer its valenced states subjectively, qualitatively, and selectively — which means that there must be something it is like to be such a system.

---

12 For readers wanting more technical detail: this is because such scalarizations obscure hard viability constraints and context-sensitive trade-offs; each need must be treated as a distinct (non-fungible) category in the context of expected outcomes. An organism (or artificial agent) must of course ultimately arrive at a single criterion for selecting among competing action policies (in our framework, this is expected 'free energy'), but that scalar quantity is computed over a model that keeps the needs distinct and ensures that each one of them is met in its own right.

If that is so, it should be possible to engineer a system that feels.

Seth (in press), in his recent comprehensive critique of computational functionalism in favour of biological naturalism with respect to consciousness, has argued that affective consciousness is 'grounded in active inference, cybernetics, autopoiesis, and the free energy principle' — all of which are substrate-independent functions. Even while defending biological naturalism, therefore, Seth concedes that 'physically embodied robots could have energetic and structural integrity requirements entailing real though non-biological forms of homeostasis and allostasis (but not autopoiesis)',[13] which in turn, he accepts, would render them 'artificially alive' and would instantiate 'real artificial consciousness'.

The AI system that our research team has been developing over the past few years is grounded in just these functional principles (in active inference, cybernetics, etc.), although in many ways it is not so different from any artificial agent that learns to play a computer game. Our agent's 'survival' depends upon its ability to act efficiently in its (virtual) environment in such a way as to meet three competing, categorically distinctive needs through non-biological forms of homeostasis and allostasis, driven by a sophisticated form of active inference grounded in the free energy principle (see Hodson *et al.*, 2025, for an early iteration of this computational architecture). Our system is continuously trying to reduce uncertainty — i.e. increase confidence or 'precision' (reduce the spread of the distribution of possible states, given its observations) — in its predictions about how to meet its existential needs through the exploitation of resources that can be found in its environment, and thereby to stay ‘alive’. This agent is genuinely agentic, in the way that Seth (in press) points out most current so-called ‘agentic AI’ is not, since it possesses what he calls ‘*endogenous* goals’. It is designed to have what Seth calls ‘skin in the game’; to ‘give a damn’ (Haugeland, 1998). This is a fundamental difference between our agent and an LLM, for example: LLMs lack existential stakes. An LLM’s ‘hallucinated’ solution has no consequence for its survival. In contrast, if our agent hallucinates incorrectly (makes a wrong inference about meeting its needs), it moves toward its own oblivion — which is registered in the form of distressing feelings.

---

[13] See Friston (2013) for experimental proof that autopoiesis, too, can be engineered artificially. For later work along these lines, see Da Costa *et al.* (2021) and Ramstead *et al.* (2023).

In its current iteration, the principal functional mechanism that drives our agent’s decision making (its ‘choices’) and its actions (its ‘voluntary behaviour’) is the efficient prioritization of its homeostatic needs in light of unfolding opportunities that it encounters and expects to encounter (on the basis of its past experience) in an ever-changing and therefore always uncertain environment. To this end, the agent palpates (i.e. senses through active exploration) and adjusts[14] its uncertainty — its confidence levels — in its current (and potential future) action policies, because fluctuations in its confidence about these policies is tethered to inferences that it makes about its underlying need states, resulting in their prioritization. In other words, the agent constantly asks itself (in abstract ways, which we anthropomorphize here): ‘*how confident am I that this is my most salient need currently, and that this is the action policy which is most likely to satisfy it?*’

In this set-up, the need that is prioritized is the one which the agent infers is the most salient (by comparing it with the current state of its other needs and by considering the opportunities it detects currently in its environment, alongside its likely future needs and opportunities): the one that most increases its confidence in its guiding hypothesis to the effect that *it will survive* (i.e. that minimizes its expected free energy). Then it selects an action policy as to how best it can satisfy the prioritized need, and it palpates its confidence in this policy as it is executed, by comparing the variance in what actually happens (both internally and externally) with what it expected would happen. On this basis, the agent adjusts its action policies on the fly, in relation to the valenced state of the currently prioritized category (i.e. quality) of need.

---

[14] Technically: this adjustment is not due to the agent’s representing its uncertainty; it is a response to new observations. Also, in its current instantiation, our agent’s uncertainty is not itself an observation; it is a property of beliefs that (mainly) drives information-seeking (in the technical sense that the word is used in this paper, ‘beliefs’ are probability distributions).

At the same time, it deploys stereotyped action policies — with fixed confidence levels — in relation to its other needs. Then it constantly re-evaluates which need should be prioritized next.

In the biological equivalent of this functional set-up, and certainly in the human one, the prioritized need is *felt*, and thereby drives the organism's cognitive choices and voluntary behaviour. For example, the organism frantically seeks oxygen because it feels that it is suffocating. In this set-up, feelings have causal power; they actually *do* something.

How could we know if (or when) this happens in our artificial agent?

### *2.3. Verification of consciousness?*

Our agent is driven by mere computations, and in many senses exceptionally simple ones. But, under functionalist assumptions, the same applies to the brain. The brain's biological effectors are driven by digital spike trains, which are likewise driven by its receptors, which are analogically modulated (via the reticular activating system palpating and adjusting post-synaptic gain) by its phenotypic generative model, which is grounded in an innate prior preference distribution — a distribution over the (homeostatic) states which it prefers to maintain. All this computational work serves a singular purpose, in the case of our agent no less than that of the brain: it ensures that the system, and the body it regulates, remains within its preferred states (so long as the updating of the model is accurate enough, and the agent is not overly constrained in its actions).

It is the fluctuating *state* of such a system's confidence as to whether it is indeed achieving likely survival, it seems to us, that is the physical instantiation of its affect. However, in order to *experience* such states, you have to *be* the system.

This is the problem of other minds: we can only *observe*, directly, our own subjective states. We can only *infer*, indirectly, the states of other minds, including artificial ones. The problem we are addressing in this paper therefore obliges us to develop methods for inferring the phenomenal states of our AI system, rather than observing them directly.

We can find no better way forwards, then, than to seek reasonable consensus from our peers, in advance, as to which methods we should use — when the time comes — to attempt to test the hypothesis that our agent *feels* deviations from and returns toward its preferred states, and that its choices and actions are affected causally by these felt states. Then we want to understand the functional mechanism by which it does so, and to compare it to that of vertebrate (and perhaps even simpler) brains. If we cannot reach reasonable consensus about such methods of verification, in advance, there is a significant risk of shifting goalposts (of our peers not accepting the implications when our agent passes what were previously considered to be critical tests).

We turn now to a consideration of one such possible method of verification.

## 3.

At a conference on the topic of 'How to Study and Understand Non-Human Consciousness' held in Kathmandu in 2024 — which was attended by many leading researchers in the field of animal consciousness as well as researchers in artificial consciousness[15] — the assembled experts were invited to select by straw poll the single test for consciousness that they thought provides the most compelling evidence for the inference that a non-human agent is conscious. An 'hedonic place preference' test won the vote. This was just an informal poll and it does not by any means imply that this is the *best* test for consciousness. However, given the affective nature of our AI system, it seems to be a good place to start (we will soon see that this test may in fact lead to a dead end, from an experimental point of view — but an enlightening one).

### *3.1. Biological hedonic place preference*

Conditioned place preference behaviour — in general, not the hedonically conditioned variety — is achieved by reliably meeting an animal's phenotypic survival needs (its nutritional needs, for example) in a particular place, with the result that it spends larger amounts of time in that location. This behaviour can be explained without invoking consciousness, as behaviourists explained all operant conditioning (although operant conditioning need not exclude the possibility of consciousness). In relation to what was said in Part 2 of this paper, conditioned place preference behaviour — although learned — can readily be attributed to innate reflexes and instincts: the animal is evolutionarily pre-programmed to follow nutritional gradients, so it does so repeatedly, and the repeated behaviour leads to the conditioning.

[15] The invited participants were Konstantin Anokhin, Pavel Balaban, Tim Bayne, Jonathan Birch, Heather Browning, Nicola Clayton, Clara Colombatto, Martin Giurfa, Graziano Fiorito, Nicholas Humphrey, Eva Jablonka, Masanori Kohda, Matthew Larkum, Paul Manger, Jennifer Mather, Bjorn Merker, Andreas Nieder, Tenzin Priyadarshi, Murray Shanahan, Henry Shevlin, Anindya Sinha, Anna Smirnova, Mark Solms, Thomas Suddendorf, Michael Tye, Giorgio Vallortigara, and Walter Veit.

The situation becomes more interesting when place preference behaviour is conditioned not by delivering substances that meet innate survival needs (food, in our example) but by novel ones that produce pleasurable feelings, despite lacking phenotypic survival value (substances like cocaine, amphetamines, opiates, and nicotine). This is *hedonic* place preference behaviour, the conditioning of which provides support for the inference that the animal so conditioned experiences feelings.

Hedonic place preference behaviour has been demonstrated repeatedly even in creatures as simple as zebrafish (Mathur, Lau and Guo, 2011). If they failed to display it, a testable prediction arising from the hypothesis that they possess feelings would have been falsified (although the lack of such behaviour would still be open to various interpretations, requiring further experiments). Why would one hypothesize that zebrafish have feelings in the first place? The answer is: because they — like all vertebrates — possess different brain circuits for hunger-generated versus taste-generated feeding behaviour (Leng, 2018), the latter of which is associated with reportable pleasure in humans. Thus, for example, we frequently eat chocolate not because we lack sugar supplies but because it tastes good (even when we do not lack sugar supplies). The same applies to cocaine, opiates, alcohol, etc. — people imbibe such substances because they like the feelings they produce.

It is easy to see why this test might inspire confidence in our animal-consciousness colleagues. Zebrafish are not pre-programmed by natural selection to seek cocaine, etc. Why, then, do they seek it after we introduce them to it? The intuitively reasonable answer is: because they *like* or *want* the feeling it produces; regardless of the fact that cocaine doesn't satisfy any nutritional need, it generates pleasurable feelings.

To be clear: there is nothing special about *place* in the hedonic place preference test. Place preference behaviour merely reflects an underlying hedonic preference (the fish are conditioned to associate a pleasurable feeling with the place in which the substance producing that feeling was administered). The feeling could just as well have been associated with a sound.

How might we set about developing an analogous test for our artificial agent? The answer is: easily — and enlighteningly so.[16]

### *3.2. Artificial hedonic place preference*

In this section, we explain how and why our agent may be expected to display hedonic place preference behaviour, and we argue that the mechanism by which it would do so can be seen as a functional basis for consciousness. Then we reflect on the hedonic place preference test, arguing that it isn't the test itself that is decisive; an inspection of the agent's underlying mechanisms is more telling.

All it takes for our agent to display the requisite behaviour is for it to have sensory inputs which are the same as those that previously (reliably) did lead it to being in a preferred state, but which now do not. In our set-up it must also have a 'Markov blanket',[17] which sequesters the inferential processing of the agent from its real bodily conditions and from the real world outside, and it must infer its movement towards

[16] To avoid confusion: in what follows we describe only the minimal instantiation required for the hedonic place preference test (endogenous needs, Markov-blanketed partial observability, and policy selection in relation to expected outcomes). Our actual, more elaborate, system instantiates the same functional principles, but the present argument does not depend upon its full engineering detail.

[17] A Markov blanket separates the internal states of a system from its surrounding external states, which are not part of the system, and it renders them conditionally independent of each other. The external states can influence the ('sensory' states of the) blanket only, which in turn can influence internal states; and the internal states can influence the ('active' states of the) blanket only, which in turn can influence external states. Thus, the internal states of a blanketed system have no *direct* causal interaction with the external world.

It is important to recognize, in this respect, that both the environment and the body are 'external' to the system; i.e. the blanket's sensory states include both exteroceptive and interoceptive states. What is internal to the system is only its *model* of the objective environment and body.

For readers wanting more technical detail: in our implementation, the blanket is instantiated very literally at the external/internal interface. The environment and body simulators maintain the external (latent) state variables, and the agent has no direct access to these variables. Instead, at each time-step, the simulators expose to the agent only a restricted observation vector (the sensory states of the blanket), generated from the latent state by an observation mapping; and the agent returns to the simulators only an action vector (the active states of the blanket), which is the sole channel by which it can influence the latent dynamics. The agent's internal states consist in its generative model and posterior beliefs (including confidence values in candidate policies), updated exclusively on the basis of observation history and prior preferences. Formally, therefore, conditional on blanket states, internal and external states are conditionally independent in the standard sense. This is what makes Oracle-like access to reality impossible for the agent, and it is precisely what makes systematic mis-inference possible when low-bandwidth sensory cues become misleading proxies for genuine need satisfaction.

its preferred state based on the (incorrect) belief that it has something of genuine ‘goodness’ about it. On this basis alone, we predict: our agent would show hedonic place preference behaviour.

This is bound to disappoint some readers. They will object — if it works — that the hedonic place preference test won’t have demonstrated that our agent possesses feelings; it will have demonstrated only that it is possible to hijack its ‘reward’ signal — a function that *was* pre-programmed in the first place. There is nothing special or magical about it. They might add that our agent, in the situation we have just described, would *behave* as if it had feelings, but there is no good reason to believe that it actually *has* them (remember the shifting goalposts).

Our response to these objections would be that, due to the problem of other minds, as mentioned already, we can only infer conscious states from observed behaviour. Therefore, *the only way to decide whether this particular behaviour is caused by a conscious state is to ask whether the mechanism by which it was produced may reasonably be considered to be a functional basis of consciousness*. We can see no other way. Human addictive behaviour may be described as a hijacked ‘reward’ signal, too. It is not reasonable to require that the behaviour in question should emerge in the absence of *any* causal mechanism, and it is equally unreasonable to require that the causal mechanism in question should not be the product of *any kind* of program. To require these things would be equivalent to believing that consciousness in humans has no physical mechanism and that it is not the product of any evolutionarily determined programming. In other words, it would require consciousness to be something spooky, something outside the purview of science.

However, in the case of our agent (and that of human beings), we believe that the functional mechanism by which hedonic place preference behaviour is displayed would not be merely that of objective reward hijacking; there is a subjective aspect too. So, let’s examine carefully the mechanism by which hedonic place preference behaviour might actually be produced in our artificial agent.

### *3.3. Interpretation of putative results*

(1) The agent would gravitate towards a location in its environment where it encountered sensory signals that had previously predicted a preferred state, because the presence of such signals would lead it to infer that an opportunity has arisen for it to reduce need X.

(2) This would lead it to prioritize need X while in or near that location, on the basis of the mistaken inference that X is currently its most salient need (since, on this occasion, the sensory signals do not in fact predict this preferred state).
(3) Such a mistake can occur only because there is (within the agent's functional architecture) a distinction between its inferences about its needs and the actual state of its body. This distinction is crucial, because the agent's inferences about its needs are *subjective* to the agent.
(4) What distinguishes the agent's subjective states from the objective reality is the presence of a Markov blanket, which partitions the agent's *model* of its body and of its environment from these things in themselves. In this functional set-up, the agent has access to the (sensory and active) states of its blanket only, never to reality itself, and so it can only model reality, it can never know it directly. On the basis of the model, the agent *infers* the most likely cause of its current sensory signals.
(5) What determines our agent's behaviour, then, given its guiding design principle, is its fluctuating *uncertainty* about its inferences concerning the current objective state of affairs. In other words, what is 'pre-programmed' in our agent's functional architecture is precisely *its capacity to use uncertainty as the primary determinant of action selection when it finds itself in unpredicted situations*.
(6) This includes uncertainty about its currently most salient need. Such needs, as stated above, are *valenced* and *qualified* variables, when considered from the viewpoint of the agent (an agent which is, in this case, actually equipped with a viewpoint).
(7) It is important to recognize that the agent's ongoing choices are *individualized*; that is, they are determined as much by its pre-programming as by each individual agent's past experiences of utilizing that programming to deal with novel situations, the adequate responses to which were not pre-programmed.

Such an agent may be described *objectively* — from the outside — as running deterministically, although it has imperfect information about its state and that of the world. But there is also — necessarily — a *subjective* non-determinism; the agent makes inferences from its own, internal perspective, and it is uncertain about those inferences. On this specific basis, our agent could have subjective experiences.

But it is important to add that these would include experiences of a notable kind, namely experiences of choice and control. This introduces the topic of free will, or at least the feeling of free will. Here we are equating 'subjective non-determinism' with deliberate choice, recognizing fully that many neuroscientists and philosophers argue that such choice — free will — is illusory, while the experience of it is not.[18] The problem of other minds is in this respect not the obstacle; it is the solution. The agent infers the state of its body and world and then feels, through the optimization of its uncertainty, that it is having an effect upon them. It is indeed having an effect upon them, but in a Markov-blanketed system with the affective functionality that we have outlined, this effect can be described both objectively (as the maintenance of a pre-programmed preference distribution) and subjectively (as the optimization of categories of feeling). These feelings are the modulation by the agent of its uncertainty about the current state of its body and the world; they are its state of uncertainty about what it might do to reduce its currently prioritized need.

If the agent had Oracle-like access to the true state of itself in the world, including the workings of its mind, there would be no place for subjectivity. There would be no room for the feeling of free will, either, since the agent would just see things ticking along deterministically. It is only because of our agent's subjective *uncertainty* — brought about by the Markov partitioning — and the valenced *quality* of its needs — brought about by the non-fungible categorization of its free energy — that both the fact of subjectivity and the feeling of free will (of subjectively selected actions) might come about.

The crucial test here is one of mechanism, not of behaviour. Therefore, one could go a step further than the hedonic place preference test, or a step simpler. If an agent with the specific functionality and architecture we have described is in a state where it can take an action which it infers will lead it closer to its preferred state, it should feel good. No complicated place preference experiment is required. Consider the example of voluntary behaviour modulated by suffocation alarm, given above. It is not the behaviour itself that is convincing (this is something very easily programmed into the system), but the fact that the functional mechanism by which the agent produces the behaviour implicates a subjective reality which only the agent has access to.

---

[18] Our own position is close to Metzinger's (2009) view that subjectivity depends on epistemic opacity in the self-model, but we provide a concrete mechanism through which such opacity gives rise to the phenomenology of agency.

So, it seems that we can create artificial agents which exhibit hedonic place preference behaviour because, under uncertainty, they can falsely infer — like zebrafish and humans — that consuming an hedonically valenced and qualified (but non-nutritive) substance will move them toward a preferred state. Such agents may plausibly be said to feel something. However, there is no magic in this behaviour *per se*, because even a much simpler situation — one in which the agent is merely perturbed away from its preferred state and must infer its condition under uncertainty — may likewise be said to involve feeling.

Both situations deal with the necessary sequestering of the uncertainty-minimizing-self from the objective-body-in-the-world. However, the first of them produces an experimentally controlled behaviour that we can describe more obviously as being affectively driven and as being determined by the previous experiences of each individual agent, and which therefore cannot be described as innate or pre-programmed. In both cases, however, the agent is in fact acting deterministically to reduce its expected free energy (the computer program runs the same way, every single time it is started); but in the hedonic place preference test it becomes more readily apparent to researchers that such systems have properties which can lead them to possess false models. The agent feels itself getting closer to its preferred state, but over time it will realize that its needs are not actually being met — when they increase to an extent that they overcome the hijacking of its low-bandwidth sensory observations.

A zombie agent with no felt uncertainty would detect the lack of nutrition in the 'cocaine' location, and then simply stop going there; it would prioritize the objective variable — nutrition — immediately. Conversely, one could envisage a zombie that did pass the hedonic place preference test but which had no uncertainty and no subjective experience, simply by having a faulty sensor which conveyed that the cocaine was nutritious. Such a zombie would soon run out of fuel. Thus, the passing of the test alone is not sufficient to tell us anything concrete about an agent's subjectivity.

### *3.4. Some additional points*

Why the low bandwidth? Why do we *not* have Oracle-like access to the true state of ourselves in the world? If our senses possessed higher fidelity, we would be less easily tricked by non-nutritional hedonic substances. We might be able to tell precisely what is happening at the level of (for example) our digestive system, our blood glucose levels, and the metabolizing of the nutrients we are taking in. Unfortunately, however, there are significant physical limits to how high such fidelities can be. The human brain, for example, at around 1.5 kg, must, through about 10 million sensory neurons, infer the state of the rest of its (approximately) 70 kg of mass containing around 30 trillion cells, and via them the state of the outside world, and thereby predict constantly what the outcome of its actions will be. Moreover, evolution has not maximized sensory fidelity since there is a trade-off between accuracy and complexity. Maintaining a simpler, lower bandwidth model is advantageous, since it is computationally more efficient and therefore metabolically cheaper, while still being sufficiently accurate for survival. Conscious feeling is evidently a consequence, therefore, not of a miraculous evolutionary success, but of our necessarily paltry access to the reality of our bodies in the world.

The same applies to conscious *perception* and *cognition* — i.e. to predictive work concerning the outside world performed in the service of each agent's endogenous needs. As is the case with zebrafish and human beings, our agent models (subjectively) only limited aspects of the external world, and it palpates and adjusts its uncertainty in only those sensory and active states of its blanket and those internal states of its generative model that are currently salient — i.e. relevant for satisfying a currently prioritized need.

The transition from affective to perceptual/cognitive qualia presumably began with what Panksepp (1998) calls 'sensory' affects, such as feelings of pain, disgust, and fright. These sensory qualities are felt, by definition. The nature of the further transition to affectless sensory qualia (if indeed they exist,[19] such as the colour blue and the sound of a middle C) is by no means obvious, beyond the fact that the latter are contingent upon — and contextualize — the former ('I feel like this *about that*'). However, the additional empirical fact that perfectly predictable sensory stimuli fade rapidly from consciousness — such as happens with a solid vertical line placed directly before the eyes and stabilized in relation to them (Riggs and Ratliff, 1951; Ditchburn and Ginsborg, 1952) — suggests that felt *uncertainty* is the common currency of both conscious perception and affect.

---

[19] See Campbell (2000), Cleeremans and Tallon-Baudry (2022).

## 4. Conclusion

A computational architecture which exhibits the crucial mechanisms and behaviours we have described in this paper is relatively simple. The important feature is to create the partitioning within the programming of its *modus operandi* for survival. An agent with such programming has access to its (in our case, simulated) bodily states and to those of the world via its Markov blanket only; so, when it receives sensory signals which trick it into mistakenly believing that its chosen actions are heading it towards a preferred state, it will indeed have the subjective feelings that are associated with that optimistic belief. Therefore, it will keep returning to that place. From the perspective of the outside observer, this is as simple as the agent's reward signal being hijacked. It is only from the perspective of the agent itself, doing the palpating of its uncertainty, that we can reasonably infer the presence of subjective feeling.

To be clear: what is crucial to us is the mechanism by which our agent might pass the hedonic place preference test, not the passing of the test itself. That is why we have concluded that one could go a step further and remove the trickery altogether, and that there is in fact no more magic to passing this test than to an even simpler arrangement, where an agent is gently pushed away from its preferred state and it has to infer — through its uncertainty — the current state of its body in the world.

What is ultimately at stake, it seems, is the functional-mechanistic justification for attributing a point of view to the agent in the first place. In our opinion, this attribution is by no means justified in the case of most AIs, but it does appear to be justified in the case of a blanketed agent with the itinerant affective dynamics that we are developing. And so, once again, we reach the impasse of other minds, but perhaps with a newfound respect for agents which feel their way through an ultimately deterministic world, believing that they possess agency, because they feel their own uncertainty and act accordingly.

*Acknowledgments*

The research described in this paper was supported by donor funding from the Oppenheimer Memorial Trust (UCT, Neuroscience Institute, internal fund number 475201 NSI1006) and Conscium, Ltd. (UCT, Dept. of Psychology, cost centre PSY526 fund number 428430). The authors wish also to acknowledge the intellectual contribution of Ryan

Smith to this research, although he bears no responsibility for the conclusions we report here.

## References

Andrews, K., Birch, J., Sebo, J. & Sims, T. (2024) *Background to the New York Declaration on Animal Consciousness*, [Online], nydeclaration.com.

Barrett, L.F. (2017) *How Emotions are Made*, New York: Houghton Mifflin Harcourt.

Campbell, N. (2000) Physicalism, qualia inversion, and affective states, *Synthese*, **124**, pp. 239–255.

Chalmers, D. (1995) Facing up to the problem of consciousness, *Journal of Consciousness Studies*, **2** (3), pp. 200–219.

Cleeremans, A. & Tallon-Baudry, C. (2022) Consciousness matters: Phenomenal experience has functional value, *Neuroscience of Consciousness*, **2022** (1), niac007. doi: 10.1093/nc/niac007

Da Costa, L., Friston, K., Heins, C. & Pavliotis, G. (2021) Bayesian mechanics for stationary processes, *Proceedings of the Royal Society*, A.47720210518. doi: 10.1098/rspa.2021.0518

Damasio, A. (2010) *Self Comes to Mind*, New York: Pantheon.

Ditchburn, R. & Ginsborg, B. (1952) Vision with a stabilized retinal image, *Nature*, **170**, pp. 36–37. doi: 10.1038/170036a0

Friston, K. (2013) Life as we know it, *Journal of the Royal Society Interface*, **10**, 20130475, doi: 10.1098/rsif.2013.0475

Haugeland, J. (1998) *Having Thought: Essays in the Metaphysics of Mind*, Cambridge, MA: MIT Press.

Hodson, R., Grimbly, S.J., Smith, R., Shock, J., Solms, M., Boonstra, E., van Hoof, C., Hakimi, N., Bassett, B. & Rosman, B. (2025) Sophisticated learning: A novel algorithm for active learning during model-based planning, *SSRN*, preprint, [Online], http://dx.doi.org/10.2139/ssrn.5392003.

Jackson, F. (1982) Epiphenomenal qualia, *Philosophical Quarterly*, **32**, pp. 127–136. doi: 10.2307/2960077

Koch, C. (2004) *The Quest for Consciousness: A Neurobiological Approach*, Englewood, CO: Roberts and Company.

Leng, G. (2018) *The Heart of the Brain: The Hypothalamus and Its Hormones*, Cambridge, MA: MIT Press.

Mathur, P., Lau, B. & Guo, S. (2011) Conditioned place preference behavior in zebrafish, *Nature Protocols*, **6**, pp. 338–345. doi: 10.1038/nprot.2010.201

Merker, B. (2007) Consciousness without cerebral cortex: A challenge for neuroscience and medicine, *Behavioral and Brain Sciences*, **30**, pp. 63–68. doi: 10.1017/S0140525X07000891

Metzinger, T. (2009) *The Ego Tunnel*, New York: Basic Books.

Mørch, H.H. (2017) The evolutionary argument for phenomenal powers, *Philosophical Perspectives*, **31**, pp. 293–316.

Mørch, H.H. (2020) The phenomenal powers view and the meta-problem of consciousness, *Journal of Consciousness Studies*, **27** (5–6), pp. 131–142.

Nagel, T. (1974) What is it like to be a bat?, *Philosophical Review*, **83**, pp. 435–450. doi: 10.2307/2183914

Nagel, T. (2021) Comment on *The Hidden Spring*, *Journal of Consciousness Studies*, **28** (11–12), pp. 203–209.

Nagel, T. (2023) Psychophysical monism as an ideal, *Keynote lecture at the Association for the Scientific Study of Consciousness Meeting*, New York.

Panksepp, J. (1998) *Affective Neuroscience: The Foundations of Human and Animal Emotions*, New York: Oxford University Press.

Ramstead, M., Sakthivadivel, D., Heins, C., Koudahl, M., Millidge, B., Da Costa, L., Klein, B. & Friston, K.J. (2023) On Bayesian mechanics: A physics of and by beliefs, *Interface Focus*, 1320220029. doi: 10.1098/rsfs.2022.0029

Riggs, L. & Ratliff, F. (1951) Visual acuity and the normal tremor of the eyes, *Science*, **114**, pp. 17–18. doi: 10.1126/science.114.2949.17

Seth, A. (2021) *Being You: A New Science of Consciousness*. London: Faber & Faber.

Seth, A. (in press) Conscious, artificial intelligence and biological naturalism, *Behavioral & Brain Sciences*. doi: 10.1017/S0140525X25000032

Solms, M. (2019) The hard problem of consciousness and the free energy principle, *Frontiers in Psychology*, **10**, art. 2714. doi: 10.3389/fpsyg.2018.02714

Solms, M. (2021) *The Hidden Spring: A Journey to the Source of Consciousness*, London: Profile Books.

Solms, M. & Friston, K. (2018) How and why consciousness arises: Some considerations from physics and physiology, *Journal of Consciousness Studies*, **25** (5–6), pp. 202–238.

Thorndike, E. (1911) *Animal Intelligence*, New York: Macmillan.

Zeman, A. (2001) Consciousness, *Brain*, **124**, pp. 1263–1289.